\documentclass[
]{ceurart}

\begin{document}

\copyrightyear{2021}
\copyrightclause{Copyright for this paper by its authors.\\
  Use permitted under Creative Commons License Attribution 4.0
  International (CC BY 4.0).}

\conference{In A. Martin, K. Hinkelmann, H.-G. Fill, A. Gerber, D. Lenat, R. Stolle, F. van Harmelen (Eds.), 
Proceedings of the AAAI 2021 Spring Symposium on Combining Machine Learning and knowledge Engineering (AAAI-MAKE 2021) - 
Stanford University, Palo Alto, California, USA, March 22-24, 2021.}

\title{"Is depression related to cannabis?": A knowledge-infused model for Entity and Relation Extraction with Limited Supervision}

\author[1]{Kaushik Roy}
\ead{kaushikr@email.sc.edu}
\address[1]{Artificial Intelligence Institute, University of South Carolina, Columbia}
\author[1]{Usha Lokala}
\ead{Nlokala@email.sc.edu}
\author[1]{Vedant Khandelwal}
\ead{vedant@mailbox.sc.edu}
\author[1]{Amit Sheth}
\ead{amit@sc.edu}

\begin{abstract}
With strong marketing advocacy of the benefits of cannabis use for improved mental health, cannabis legalization is a priority among legislators. However, preliminary scientific research does not conclusively associate cannabis with improved mental health. In this study, we explore the relationship between depression and consumption of cannabis in a targeted social media corpus involving personal use of cannabis with the intent to derive its potential mental health benefit. We use tweets that contain an association among three categories annotated by domain experts - Reason, Effect, and Addiction. The state-of-the-art Natural Langauge Processing techniques fall short in extracting these relationships between cannabis phrases and the depression indicators. We seek to address the limitation by using domain knowledge; specifically, the Drug Abuse Ontology for addiction augmented with Diagnostic and Statistical Manual of Mental Disorders lexicons for mental health.  Because of the lack of annotations due to the limited availability of the domain experts' time, we use supervised contrastive learning in conjunction with GPT-3 trained on a vast corpus to achieve improved performance even with limited supervision. Experimental results show that our method can significantly extract cannabis-depression relationships better than the state-of-the-art relation extractor. High-quality annotations can be provided using a nearest neighbor approach using the learned representations that can be used by the scientific community to understand the association between cannabis and depression better.
\end{abstract}


\maketitle
\section{Introduction}
\label{sec:intro}
Many states in the US have legalized the medical use of cannabis for therapeutic relief in those affected by Mental Illness \cite{Hanson2014-jh}
\cite{Room2014-fk, Volkow2014-jl}. The use of cannabis for depression, however, is not authorized yet \cite{Bridgeman2017-er}. Depression is ubiquitous among the US population, and some even use cannabis to self treat their depression\cite{Young2008-jg}\cite{Lankenau2018-rh}. Therefore, scientific research that can help understand the association between depression and cannabis consumption is of the utmost need given the fast increasing cases of depression in the US and consequent cannabis consumption \cite{Keyes2016-wd}. \\
\indent Twitter can provide crucial contextual knowledge in understanding the usage patterns of cannabis consumption concerning depression \cite{Corazza2013-qf, Burns2014-ic}. Conversations on social media such as Twitter provide unique insights as tweets are often unfiltered and honest in disclosing consumption patterns due to the anonymity and private space afforded to the users. For now, even with several platforms available to aid the analysis of depression concerning cannabis consumption, this understanding remains ambiguous  \cite{Cavazos-Rehg2018-zc, Daniulaityte2017-fc, Lamy2018-ou}. Still, encouragingly there is support in the literature to show that cannabis use can be potentially associated with depressive patterns. Hence, we aim to extract this association as one of three relationships annotated by domain experts: Reason, Effect, and Addiction (Table \ref{tab:table1}).

\begin{table}[pos = h!]
  \begin{center}
       \begin{tabular}{p{2cm}|p{8cm}}
 \hline
 \hline
      \textbf{Relationship} & \textbf{Tweet}\\
      \hline
      \multirow{2}{*}
      {Reason}  
      & ``-Not saying im cured, but i feel less \textcolor{red}{depressed} lately, could be my \#\textcolor{blue}{CBD oil} supplement." \\ 
      \hline
      Effect  & ``-People will \textcolor{blue}{smoke weed} and be on antidepressants. It's a clash!Weed is what is making you \textcolor{red}{depressed}." \\
      \hline
      Addiction & ``-The lack of \textcolor{blue}{weed} in my life is \textcolor{red}{depression} as hell." \\ \hline
    \end{tabular}
     \caption{cannabis-depression Tweets and their relationships. Here the text in the \textcolor{blue}{blue} and \textcolor{red}{red} represents the cannabis and depression entities respectively.}
    \label{tab:table1}
  \end{center}
  \vspace{-1.5 em}
\end{table}

\textbf{The paper studies mental health and its relationship with cannabis usage, which is a significant research problem. The study will help address several health challenges such “as the investigation of cannabis for the treatment of depression”,” as a reason for depression” or “as an addictive phenomenon that accompanies depression”.} Extracting relationships between any \textit{concepts/slang-terms/synonyms/street-names} related to `\textit{cannabis},’ and `\textit{depression},’ from text is a tough problem. This task is challenging for traditional Natural Language Processing (NLP) because of the immense variability with which tweets mentioning depression and cannabis are described. Here, we make use of the Drug Abuse Ontology (DAO) \cite{Cameron2013-qq,noauthor_undated-xp} which is a domain-specific hierarchical framework containing 315 entities (814 instances) and 31 relations defining concepts about drug-abuse. The ontology has been used in prior work to analyze the effects of cannabis \cite{Lokala2018-dm, cameron2013predose, kursuncu2018s}. DAO was augmented using Patient Health Questionnaire 9th edition (PHQ-9), Systematized Nomenclature of Medicine - Clinical Terms (SNOMED-CT), International Classificatio of Diseases 10th edition (ICD-10), Medical Subject Headings (MeSH) Terms, and Diagnostic and Statistical Manual for Mental Disorders (DSM-5) categories to infuse knowledge of mental health-related phrases in association with drugs such as cannabis \cite{Gaur2018-xv}\cite{gaur2019knowledge}. Some of the triples extracted from the DAO are as follows: \textbf{(1)} SCRA $\rightarrow$ subClassOf $\rightarrow$ Cannabis; \textbf{(2)} Cannabis\_Resin $\rightarrow$ has\_slang\_term $\rightarrow$ Kiff; \textbf{(3)} Marijuana\_Flower $\rightarrow$ type $\rightarrow$ Natural\_Cannabis. \\ 
\indent For entity and relationship extraction (RE), previous approaches generally adopt deep learning models \cite{lin2016neural}  \cite{lee2019semantic}. However, these models require a high volume of annotated data and are hence unsuitable for our setting.
Several pre-trained language representation models have recently advanced the state-of-the-art in various NLP tasks across various benchmarks \cite{maillard2019jointly,akbik2019pooled}. GPT-3 \cite{brown2020language}, BERT \cite{devlin-etal-2019-bert} are such language models \cite{lin2019bert}. Language models benefit from the abundant knowledge that they are trained on and, with minimal fine-tuning, can tremendously help in downstream tasks under limited supervision. Hence, we exploit the representation from GPT-3 and employ supervised contrastive learning to deal with limited supervision in terms of quality annotations for the data. 
We propose a knowledge-infused deep learning framework based on GPT-3 and domain-specific DAO ontology to extract entities and their relationship. Then we further enhance the utilization of limited supervision through the use of supervised contrastive learning. It is well known that deep understanding requires many examples to generalize. Metric Learning frameworks such as Siamese networks have previously shown how limited supervision can help use contrastive learning with triplet loss \cite{hoffer2015deep}. Combinatorially this method leads to an increase in the number of training examples from an order of $n$ to $nC3$, which helps with generalizability. The technique can also exploit the learned metric space representations to provide high-quality annotations over unlabeled data. Therefore, the combination of knowledge-infusion \cite{sheth2019shades, kursuncu2019knowledge, gaur2020knowledge}, pre-trained GPT-3, and supervised contrastive learning presents a very effective way to handle limited supervision. The proposed model has two modules: 
\textit{\textbf{(1) Phrase Extraction and Matching Module}}, which utilizes the DAO ontology augmented with the PHQ-9, SNOMED-CT, ICD-10, MeSH Terms, and Diagnostic and Statistical Manual for Mental Disorders (DSM-5) lexicons to map the input word sequence to the entities mention in the ontology by computing the cosine similarity between the entity names (obtained from the DAO) and every n-gram token of the input sentence. This step identifies the depression and cannabis phrase in the sentence. Distributed representation obtained from GPT-3 of the depression phrase and cannabis phrase in the sentence is used to learn the contextualized syntactic and semantic information that complement each other. 
\textit{\textbf{(2) Supervised Contrastive Learning Module}}, uses a triplet loss to learn a representation space for the cannabis and depression phrases through supervised contrastive learning. Phases with the correct relationship are trained to be closer in the learned representation space, and phrases with incorrect relationships are far apart.\newline
\textbf{Contributions}:\\
\textbf{(1)} In collaboration with domain experts who provide limited supervision on real-world data extracted from Twitter, we learn a representation space to label the relationships between cannabis and depression entities to generate a cannabis-depression relationship dataset. \\
\textbf{(2)} We propose a knowledge-infused neural model to extract cannabis/depression entities and predict the relationship between those entities. We exploited domain-specific DAO ontology, which provides better coverage in entity extraction. \\
\textbf{(3)}Further, we use GPT-3 representations in a supervised contrastive learning approach to learn a representation space for the different cannabis and depression phrase relationships due to limited supervision.\\
\textbf{(4)} We evaluated our proposed model on the real world twitter dataset. The experimental results show that our model significantly outperforms the state-of-the-art relation extraction techniques by $>$ 11\% points on the F1 score.

\subsection{Novel Contributions of the paper}
\textbf{(1)} Semantic filtering: We use DAO, DSM-5, to extract contextual phrases expressed implicitly in the user tweet, mentioning Depression and Cannabis. This is required for noise-free domain adaptation of the model, as evident in our results.\\
\textbf{(2)} We develop a weak supervision pipeline to label the remaining 7000 tweets with three relationships (Reason, Effect, Addiction).\\
\textbf{(3)} We learn a domain-specific distance metric between the phrases, leveraging pre-trained GPT-3 embeddings of the extracted phrases and their relationship, in a supervised contrastive loss training setup.\\
\textbf{(4)} 7000 tweets were annotated and evaluated using the learned metric against the expert annotation with clustering (TSNE).


\section{Related Works}

Based on the techniques and their application to health, we discuss recent existing works. Standard DL approaches based on Convolutional Neural Networks (CNN), and Long Term Short Term Memory (LSTM) networks have been proposed for RE \cite{liu2013convolution} \cite{miwa2016end,yadav2019feature}. Hybrid models that combine CNN and LSTM have also been proposed \cite{Liang2017-hy}. More recently, Graph Convolutional Neural Networks (GCN)' s have been utilized to leverage additional structural information from dependency trees towards the RE task \cite{pmlr-v97-wu19e}. \cite{guo-etal-2019-attention} guide the structure of the GCN by modifying the attention mechanism. Adversarial training has also been explored to extract entities and their relationships jointly \cite{bekoulis-etal-2018-adversarial}. Due to the variability in the specification of entities and relationships in natural language, \cite{choi2018extraction, peng2017deep} have exploited entity position information in their DL framework. Models have demonstrated state of the art in RE based on the popular BERT language model, BioBERT \cite{lee2019biobert}, SciBERT \cite{beltagy2019scibert}, and XLNet \citet{yang2019xlnet}. Task-specific adaptations of BERT have been used to enhance RE in \citet{shi2019simple} and \citet{xue2019fine}. \citet{wang2019extracting} augment the BERT model with a structured prediction layer to predict multiple relations in one pass. In all the approaches discussed so far, knowledge has not been a component of the architecture \cite{gaur2020semantics}.\\
\indent \citet{Chan2010-pg} show the importance of using knowledge to improve RE on sentences by showing an improvement of 3.9\% of F1-score incorporating knowledge in an Integer Linear Programming (ILP) formulation.\citet{Wen2018-en} use the attention weights between entities to guide traversal paths in a knowledge graph to assist RE. \citet{distiawan2019neural} use knowledge graph TransE embeddings in their approach to improving performance. Some of the other prominent work utilizing knowledge graph for relation extraction is \cite{li2019dual,zhou2019knowledge,li2019improving}.\\
\indent These methods, however, do not consider a setting in which the availability of high-quality annotated data is scarce. We use knowledge to extract relevant parts of the sentence \cite{alambo2019question, kursuncu2019modeling} and pre-trained GPT-3 representations trained over a massive corpus in conjunction with supervised contrastive learning to achieve a high degree of sample efficiency with limited supervision.

\section{Our Approach}
\subsection{Dataset}
The dataset we have used for our study consists of ~11,000 Tweets collected using the twitris API from Jan 2017 to Feb 2019 - determined by three substance use epidemiologists as a period of heightened Cannabis consumption. The experts annotated 3000 tweets (due to time constraints) with one of 3 relationships that they considered essential to identify: “Reason,” “Effect,” and “Addiction.” The annotation had a Cohen Kappa Agreement of 0.8. Example from each of these different relationships already shown in Table \ref{tab:table1}

\subsection{Phrase Extraction and Matching}
We exploit the domain-specific knowledge base to replace the phrases in social media text with the knowledge base concepts under this method. The Phrase Extraction and Matching are performed in several steps, which are:
\begin{itemize}
\item \textbf{Depression and cannabis Lexicon:} We have exploited the state of the art Drug Abuse Ontology (DAO) to extract various medial entities and slang terms related to cannabis and depression. We further expand the entities using entities extracted from PHQ-9, SNOMED-CT, ICD-10, MeSH Terms, and Diagnostic and Statistical Manual for Mental Disorders (DSM-5). 
\item \textbf{Extracting N-Grams from Tweets:} The N-Grams are extracted from the tweets are considered to better understand the context of the terms by taking into consideration the words around it. For example, from the tweet \textit{whole world emotionally depressed everybody needs smoke blunt to relax We living nigga}, we will obtain ngrams such as whole world, emotionally depressed, depressed everybody, need smoke, need smoke blunt, living nigga.
\item \textbf{GPT-3:}Generative Pre-Trained Transformer 3 is an autoregressive language model. We have used GPT-3 to generate the embedding of the N-Grams generated and the cannabis and depression Phrase because of the vast dataset it is trained on, which provides us the phrases' embeddings based on its global understanding.
Cosine Similarity: It is a measure of similarity between two non-zero vectors in a similar vector space. This metric is often used to get a semantic similarity between two phrase embeddings obtained in the same vector space.
\item \textbf{Phrase Matching:} We use the cosine similarity metric to understand the semantic similarity between the phrases. We have taken a threshold of 0.75 or more cosine similarity. Once the phrase obtains a similarity value more than or equals the threshold, the original N-Grams from the tweet text are replaced by the matched cannabis/depression Phrase. 
The above steps are repeated for all the tweets. For example, we would obtain \textit{emotionally depressed} as the depression phrase, whereas \textit{need smoke blunt} is found to be the cannabis phrase.

\end{itemize}
\begin{figure}[pos=t]
    \centering
    \includegraphics[width = 0.6\textwidth]{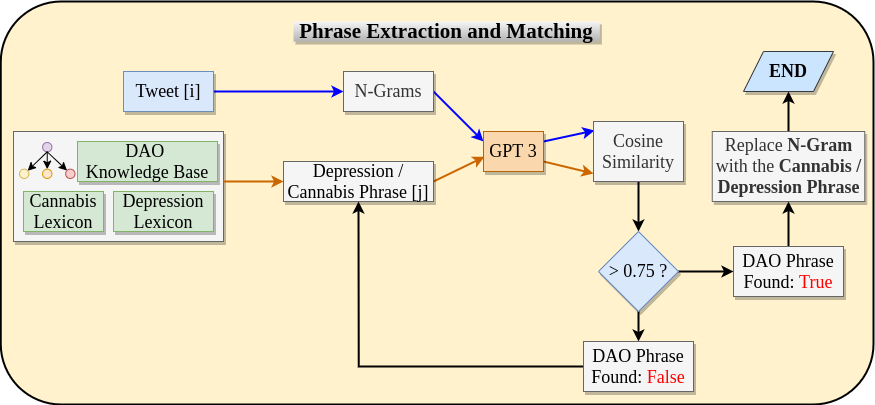}
    \caption{The Phrase extraction and Mapping Pipeline, where the DAO is used to extract GPT-3 representations of the related phrases}
    \label{fig:my_label}
\end{figure}
\subsection{Supervised Contrastive Learning}
The proposed model architecture is divided into two sub-parts: A) Transformer Block, B) Loss Function. The input tweet is first sent through a block of 12 transformers, and later the embedding is passed through a triplet loss function. The label associated with the tweet input is used to extract a complimentary sample (a tweet with the same label) and a negative sample (a tweet associated with a different label). These positive and negative samples are sent through a block of 12 transformers to obtain their embedding, which is further passed on to the triplet loss function. Under the loss, the function tries to achieve a low cosine similarity between the tweet and its negative sample as close to 0. At the same time, it tries to achieve a high cosine similarity between the tweet and its positive sample as close to 1.

    \begin{equation}
            CoSim(A,P) - CoSim(A,N) + \alpha  \leq 0
    \end{equation}
Where A is the anchor (the initial data point), P is a positive data point which is of the same class as the anchor, N is a negative data point which is of the different class as the anchor. CoSim(X, Y) is the cosine similarity between the two data points, and \( \alpha \) is the margin. For the example shown in Section 3.2, if we consider the anchor sample to be \textbf{"whole world emotionally depressed everybody needs smoke blunt relax We living nigga"}, corresponding to that positive sample is \textit{"Depressionarmy weed amp sleep I awake I depressed"} and the negative sample would be \textit{"This weird rant like weed makes anxiety depression worse Im soooo sick ppl like jus"}.

\begin{figure*}[pos=ht]
    \centering
    \includegraphics[width = \textwidth]{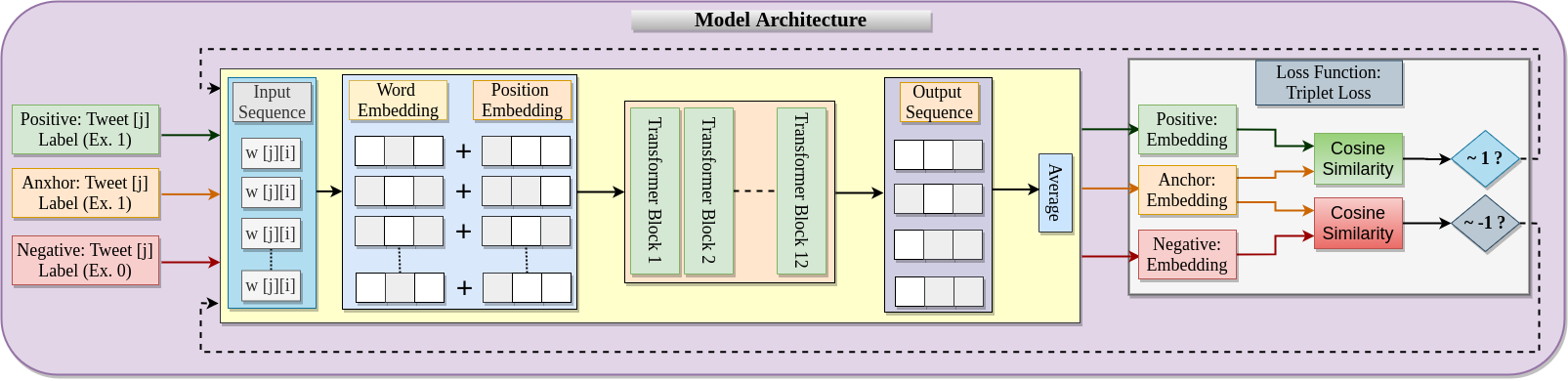}
    \caption{Supervised Constrastive Learning pipeline using Triplet Loss}
\end{figure*}

\section{Experimental Setup and Results}
In this section, we discuss the results of the task of cannabis and depression RE tasks. After that, we will also provide a technical interpretation of the results.

\subsection{Results}
The dataset utilized in our experiment is described in Section-3. We have used Precision, Recall, and F1 score as the metric to compare our proposed methodology with the state-of-the-art relation extractor. As a baseline model, we used BERT, BioBERT, and its various variations such as:
\begin{itemize}
    \item \textbf{BERT\textsubscript{PE}}: We expand the BERT as a position embedding along with the BERT embedding with the position data (the relative distance of the word with cannabis and depression entity) obtained via domain-specific knowledge resource. 
    \item \textbf{BERT\textsubscript{PE+PA}}: This consists of an additional component of position-aware mechanism along with BERT\textsubscript{PE}. 
\end{itemize}

Table-2 Summarises the performance of our model over the baselines. Our proposed methodology has outperformed all the state of the art models based on the given metrics. As compared to the worst-performing baseline, BioBERT our model achieves an F1 score with an absolute improvement of 12.32\%, and with best performing baseline BERT\textsubscript{PE+PA} it gives an improvement of 11.28\% in F1 Score. From the above comparison using contrastive learning with knowledge, infused learning can perform better in relation classification. 

\begin{table}[pos=!htbp]
    \centering
    
    \begin{tabular}{|m{9em}|m{3.5em}|m{3em}|m{3em}|}
    \hline
      Method  &  Precision & Recall & F1-Score \\
       \hline
      BERT    &  64.49  & 63.22 & 63.85 \\
       \hline
      BioBERT    &  63.97  & 62.15 & 63.06 \\
       \hline
      BERT\textsubscript{PE}    &  60.64  & 56.51 & 58.50 \\
       \hline
      BERT\textsubscript{PE+PA}    &  65.41  & 65.25 & 64.50 \\
       \hline
      \textbf{Proposed Model}   &  74.6  & 76.17 & 75.37 \\
       \hline
    \end{tabular}
    \caption{Performance comparison of our proposed model with baselines methods}
    \label{tab:Comparison-with-baselines}
\end{table}

\subsection{Ablation Study}

We have performed the ablation study by removing one component from the proposed model, evaluate its performance to understand the impact of various components. Based on the study, we found that, as we remove the contrastive loss function from our learning approach, the model performance significantly drops by 6.46 \% F1 Score, 6.53 \% Recall, and 6.4 \% Precision. The significant decrease in the model’s performance shows that generating an embedding for two samples of the same class similar and of different classes dissimilar brings in a great advantage to the training of the model. The contrastive loss function allows us to learn the representation of the same classes to be closer to each other in vector space and hence allows us in generating the representation of unlabelled data from the dataset (discussed further, later in this section )

We also observe that domain-specific knowledge resource with contextualized embedding trained over a large corpus (GPT-3) is very important. As we further remove the second component from our model, we see a total decrease of 9.01 \% F1 score, 8.92 \% Precision, and 9.11 \% recall in the proposed model’s performance. This component was majorly responsible for removing the data’s ambiguity using the phrases from human-curated domain-specific knowledge bases (such as DAO, DSM-5, SNOMED-CT, and others). Also, the contextualized embedding helped us consider the global representation of the entities present in the dataset and hence contribute to improving the model’s performance. 

\begin{table}[pos=ht]
    \centering
    
    \begin{tabular}{|m{12em}|m{7em}|m{7em}|m{7em}|}
    \hline
      Model  &  Precision & Recall & F1-Score \\
       \hline
      \textbf{Proposed Model}   &  74.6  & 76.17 & 75.37 \\
       \hline
      (-) Contrastive learning loss    &  68.2 ($\downarrow$8.5\%)  & 69.64 ($\downarrow$8.6\%) & 68.91 ($\downarrow$8.57\%) \\
       \hline
     (-) knowledge infusion    &  65.68 ($\downarrow$11.9\%) & 67.06 ($\downarrow$11.96\%) & 66.35 ($\downarrow$11.97\%) \\
       \hline
     
    \end{tabular}
    \caption{Ablation Study over the proposed model to evaluate the effect of contrastive learning loss and knowledge infusion in determining the relationship between cannabis and depression} 
    \label{tab:Ablation Study}
\end{table}

Thus, this shows that every component of the proposed model is necessary for the best performing results. 

\subsection{Cluster Analysis}
\begin{figure}[pos=!h]
    \centering
    \includegraphics[width=0.45\textwidth]{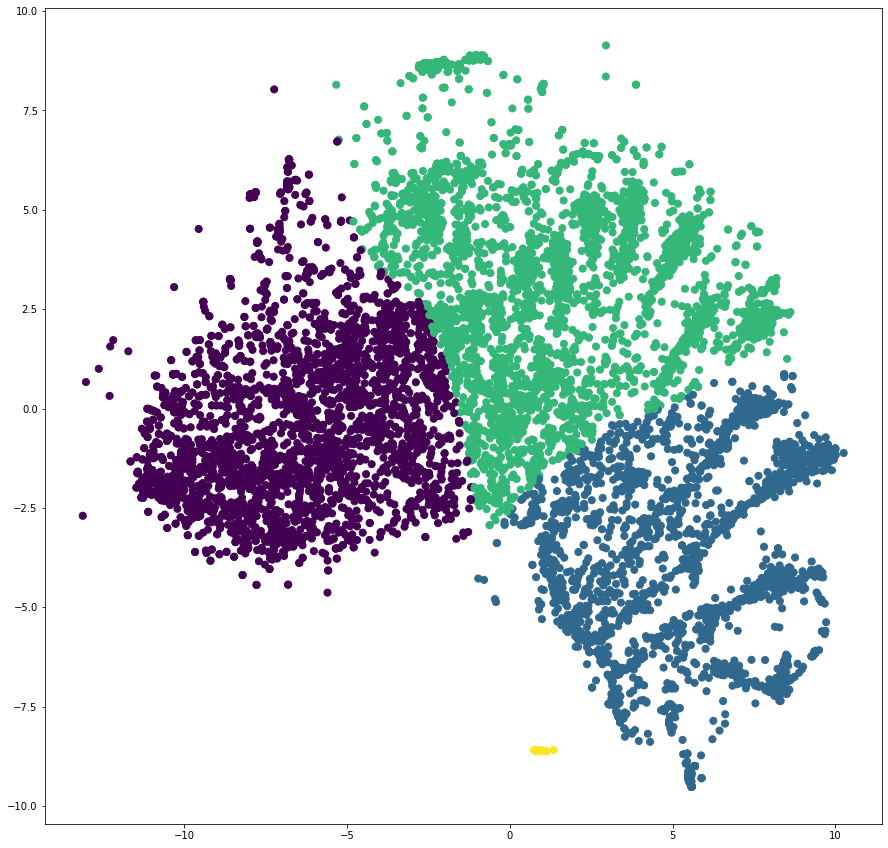}
    \caption{Shows the depression and cannabis phrases grouped according to the relationships of Reason, Effect and Addiction. Purple - Reason, Green - Effect, Blue and a little yellow below - Addiction.}
    \label{fig:my_labelClusters}
\end{figure}

After training the model, we annotate the unlabelled data in the dataset by classifying among the three relationships. We parse the unlabelled tweets from the first module to extract the phrases from the knowledge bases using contextualized embedding. Later the embeddings are pushed into the proposed model architecture to obtain a representation of the tweets. The representation is used to create a cluster of the tweet data points and determine the label of the un-labeled tweets based on the majority of the data points present. The representation of the cluster after labeling unlabelled tweets is shown in Figure \ref{fig:my_labelClusters}. Some examples from each of the cluster are as follows: 
\begin{itemize}
\item \textbf{Reason:} 1) \textit{Depressionarmy weed amp sleep I awake I depressed,} 2) \textit{mood smoke blunt except the fact I depressed, }3) \textit{weed hits ya RIGHT depression, }4) \textit{I smoked weed drank alcohol drowning sorrows away, }5) \textit{whole world emotionally depressed everybody need smoke blunt relax We livin nigga}
\item \textbf{Effect:} 1) \textit{marijuana bad marijuana makes feel depressed low mmk, }2) \textit{Unemployed stoners are the most depressed on the planet, }3) \textit{guess depression took long time discover marijuana makes VERY DEPRESSED alcohol doesnt help either, }4) \textit{This weird rant like weed makes anxiety depression worse Im soooo sick ppl like jus, }5) \textit{waking weed naps nigga feeling depressed hell} 
\item \textbf{Addiction:} 1) \textit{I feel like weed calm someone suffer depression anxiety psychosis predisposed either, }2) \textit{Small trigger warning Blaine suffers anxiety depression occasionally smoke pot, }3) \textit{need blunt accompany depression, }4) \textit{This bot crippling depression ok weed lol, }5) \textit{Violate blunt distraction possibly despair bask commitment This would never happen}
\end{itemize}

\section{Reproducibility}
From this study, we will be delivering high quality annotated dataset of 3000 tweets along with the full annotated dataset (by our method) of 11000 tweets, will be made publicly available to support research on the psychological impact of cannabis and depression use during COVID-19, as Cannabis use related to depression is seeing a rise once more. Also, the trained model will be shared for reproducibility of the results and annotation of tweets. Unfortunately, we cannot release the code used for training at this time as recently, Microsoft bought the rights to the GPT-3 model. Therefore, to use the learning method proposed in this paper, GPT-3 will need to be substituted with an alternative language model such as BERT, GPT-2, etc. \footnote{\url{https://blogs.microsoft.com/blog/2020/09/22/microsoft-teams-up-with-openai-to-exclusively-license-gpt-3-language-model/}}

\section{Conclusion}

In this study, we present a method to determine the relationship between depression and cannabis consumption. We motivate the necessity of understanding this issue due to the rapid increase in cases of depression in the US and across the world and subsequent increase in cannabis consumption. We utilize tweets to understand the relationship as tweets are typically unfiltered expressions of simple usage patterns among cannabis users who use it in association with their depressive moods or disorder. We present a knowledge aware method to determine the relationship significantly better than the state-of-the-art effectively, show the quality of the learned relationship through visualization on TSNE based clusters, and annotate the unlabeled parts of the dataset. We show by training on this new dataset (human-labeled and estimated label) that the model's prediction quality is improved. We present this high-quality dataset for utilization by the broader scientific community in better understanding the relationship between depression and cannabis consumption.

\section{Broader Impact}
Although we develop our method to handle relationship extraction between depression and cannabis consumption specifically, we generally develop a domain knowledge infused relationship extraction mechanism that uses state-of-the-art language models, few shot machine learning techniques (contrastive loss) to achieve efficient and knowledge guided extraction. We see the improved quality in the results over transformer models. We believe that for applications with real-life consequences such as these, it is crucial to infuse domain knowledge as a human would combine with language understanding obtained from language models to identify relationships efficiently. Humans typically can learn from very few examples. Motivated by this and the lack of availability of examples, we develop our relation extraction method. We hope our significantly improved results will encourage scientists to explore further the use of domain knowledge infusion in application settings that demand highly specialized domain expertise. 
\bibliography{references}

\end{document}